\documentclass[letterpaper, 10 pt, conference]{ieeeconf}  %

\IEEEoverridecommandlockouts                              %

\usepackage{graphics} %
\usepackage{epsfig} %
\usepackage{times} %
\usepackage{amsmath} %
\usepackage{amssymb}  %
\usepackage[hyphens]{url}  %
\usepackage{graphicx} %
\usepackage[utf8]{inputenc}
\usepackage[style=ieee, maxbibnames=99, citestyle=numeric-comp]{biblatex}
\usepackage{multicol}
\usepackage{booktabs}
\graphicspath{{../}{./}{./figures/}}
\usepackage{svg}
\usepackage{amsmath}
\usepackage[font=footnotesize,labelfont=bf]{caption}
\usepackage{graphicx}
\usepackage{xcolor}
\usepackage[justification=centering]{subfig}
\usepackage{amsmath}
\usepackage{amsfonts}
\usepackage{amssymb}
\usepackage{mathtools}
\usepackage{todonotes}
\let\labelindent\relax
\usepackage{enumitem}
\usepackage{comment}
\usepackage[per-mode=symbol]{siunitx}
\DeclareSIUnit\minute{min}
\usepackage{tikzpagenodes}
\usepackage{algorithm}
\usepackage[noend]{algpseudocode}

\DeclareMathOperator*{\argmax}{argmax}

\usepackage{mathrsfs}
\usepackage[bookmarks=true]{hyperref}
\hypersetup{hidelinks,colorlinks=false}

\title{\LARGE \bf
Trajectory Optimization for Adaptive \\ Informative Path Planning with Multimodal Sensing
}

\author{
    Joshua Ott\textsuperscript{\rm 1}\thanks{\textsuperscript{\rm 1}Department of Aeronautics \& Astronautics, Stanford University \newline \indent (\{joshuaott, mykel\}\!@stanford.edu).},
    Edward Balaban\textsuperscript{\rm 2}\thanks{\textsuperscript{\rm 2}NASA Ames Research Center (edward.balaban@nasa.gov).},
    and Mykel J. Kochenderfer\textsuperscript{\rm 1}
}

\begin{document}

\maketitle
\thispagestyle{empty}
\pagestyle{empty}

\begin{abstract}
We consider the problem of an autonomous agent equipped with multiple sensors, each with different sensing precision and energy costs. The agent's goal is to explore the environment and gather information subject to its resource constraints in unknown, partially observable environments. The challenge lies in reasoning about the effects of sensing and movement while respecting the agent's resource and dynamic constraints. We formulate the problem as a trajectory optimization problem and solve it using a projection-based trajectory optimization approach where the objective is to reduce the variance of the Gaussian process world belief. Our approach outperforms previous approaches in long horizon trajectories by achieving an overall variance reduction of up to 85\% and reducing the root-mean square error in the environment belief by 50\%. \textcolor{black}{This approach was developed in support of rover path planning for the NASA VIPER Mission.}
\end{abstract}

\section{Introduction}\label{sec:intro}
Adaptive informative path planning with multimodal sensing (AIPPMS) considers the problem of an agent tasked with exploring an unknown environment with finite energy resources \cite{choudhury2020adaptive}. The agent is equipped with multiple sensors each with different sensing cost and sensing precision. The agent must not only reason about where to explore but also what measurements to take at a specific location. In addition, the agent must maintain a belief about the true state of the environment since it can only partially observe the underlying state through noisy observations. To solve an AIPPMS problem, the agent must develop a flexible decision making strategy that navigates towards the desired destination, balancing the need for gathering information with utilizing the gathered information, and taking into account resource limitations along the way.

Many real-world situations involve agents equipped with multiple sensors which cannot operate concurrently due to limitations related to energy, sensor interference, or motion constraints. Consider an agent with multimodal sensing capabilities. Here, multimodal refers to the presence of numerous sensors, each with differing degrees of precision and cost; for instance, a rover equipped with an array of sensors like CCD cameras, radar, mass spectrometers, and drills. Certain sensors like the cameras consume less energy yet yield less dependable measurements since they can't completely examine the subsurface environment. In contrast, other sensors, such as the drill, consume significantly more energy but provide almost perfect measurements of the subsurface environment. This dichotomy between sensor efficiency and precision is also observable in numerous other domains, including ocean monitoring, mineral prospecting, aerial surveillance, and wildfire mapping \cite{betterton2022reinforcement, marchant2014sequential, francis2017aegis, mangold2021perseverance}.

\begin{figure}[t]
\centering
    {\includegraphics[width=1.0\columnwidth]{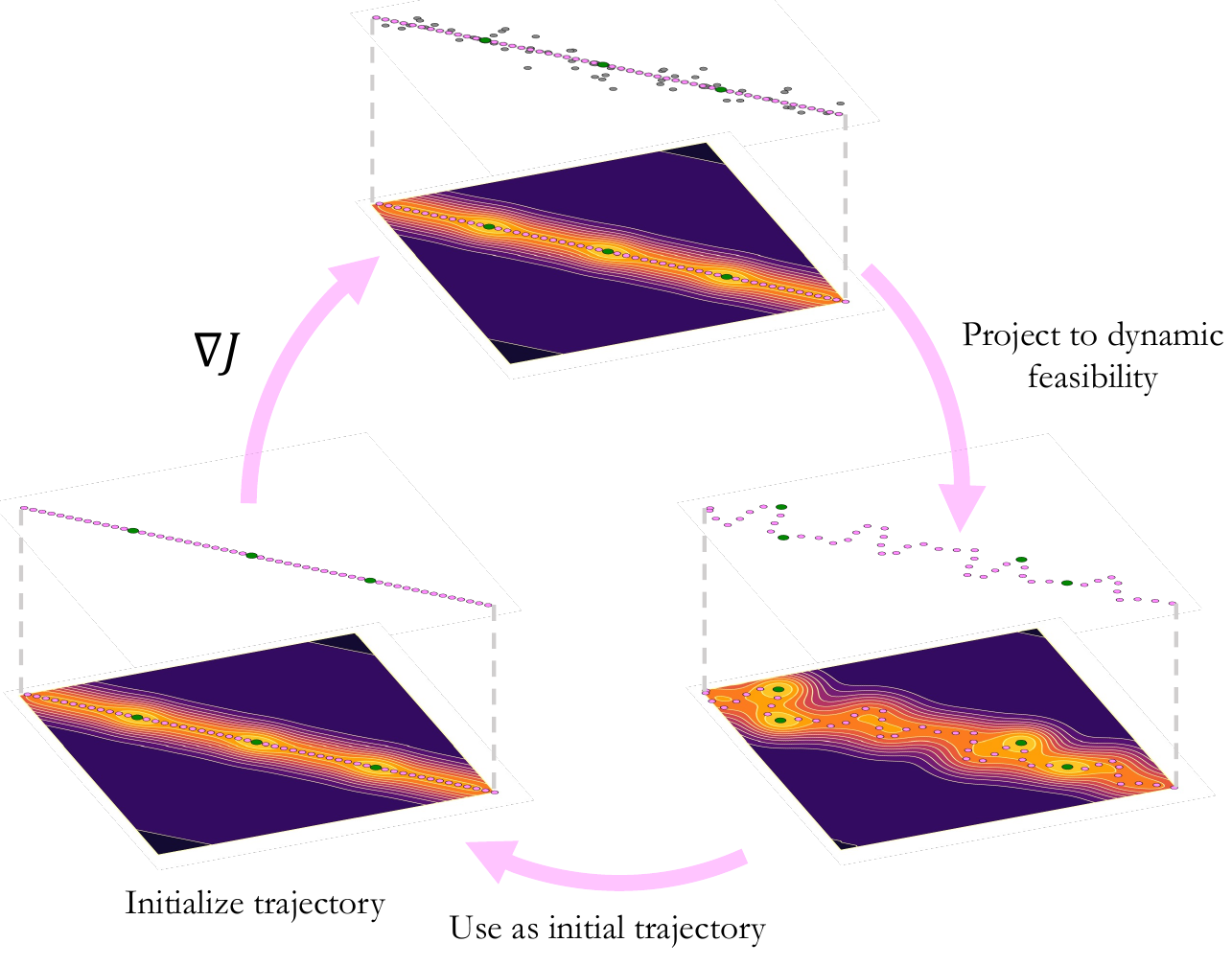}}    
  \caption{High-level overview of the projection-based trajectory optimization approach. The trajectory and sample locations are initialized. A descent direction is then computed to modify the trajectory in the direction that maximally reduces the objective function. This trajectory may violate the dynamics of the agent so the candidate trajectory is then projected back into feasibility. This process is repeated until convergence.} \label{fig:front_page_fig}
  \vspace{-7mm}
\end{figure}

The informative path planning (IPP) problem is NP-hard \cite{meliou2007nonmyopic} and involves an agent planning a path without accounting for noisy observations of the environment. The adaptive problem extends the IPP problem by seeking a policy that changes in response to received observations \cite{singh2009nonmyopic, hollinger2013active, hitz2014fully, lim2016adaptive, girdhar2016modeling}. The AIPPMS problem further extends the adaptive problem by reasoning over multiple sensing modalities with different sensing costs corresponding to sensor precision and movement costs \cite{choudhury2020adaptive}.

A variety of approaches have been proposed for both the adaptive informative path planning (AIPP) and non-adaptive IPP problems \cite{toussaint2014bayesian, popovic2020informative, duecker2021embedded, ott2022riskaware, morere2018continuous, ruckin2022adaptive}. The main focus in the majority of this previous work is on the case where the agent is equipped with a single sensor. In this work, we are interested in the case where the agent can choose between multiple sensing modalities. This extension adds additional complexity by requiring that the agent reasons about the cost-benefit trade-off associated with the different sensing modalities. \textcolor{black}{Choudhury et al. and Ott et al. both considered rollout-based partially observable Markov decision process (POMDP) solvers \cite{kaelbling1998planning, couetoux2011continuous, ott2022sequential, kochenderfer2022algorithms, kochenderfer2019algorithms}. %
Rollout-based POMDP approaches suffer from near-sightedness induced by the rollout depth. In longer trajectories, the agent may use more valuable resources at the beginning of the trajectory due to the limited rollout depth. As a result, valuable samples may be concentrated at the start of the trajectory instead of being distributed throughout the environment. To solve this problem, we propose optimizing the whole trajectory using a Gaussian process belief objective. Our holistic approach is able to better allocate valuable sensing resources across the entire trajectory and outperforms rollout-based solvers in trajectories with greater energy resources.}

The key contributions of this work are: \begin{enumerate}[leftmargin=5mm]
\item The formulation of the AIPPMS problem as a trajectory optimization problem where the objective is based on the Gaussian process world belief.
\item The introduction of a projection-based trajectory optimization method that uses Gaussian process information objectives. 
\item The extensive evaluation of our method in simulation, along with directly comparing to previous AIPPMS methods from the literature.
\item The release of our implementation as an open-source software package for use and further development by the community.\footnote[2]{\url{https://github.com/josh0tt/TO_AIPPMS}}%
\end{enumerate}

\section{Related Work}\label{sec:relatedwork}
The AIPP problem has been widely studied and a common approach is to represent the world belief as a Gaussian process. Marchant et al. formulate the AIPP problem as a POMDP and solve it using sequential Bayesian optimization through Monte Carlo tree search upper confidence bound for trees (MCTS-UCT) \cite{marchant2014sequential}. This work has been extended to modify the reward function for achieving monitoring behavior that exploits areas with high gradients \cite{morere2017sequential}, and for additional reasoning over continuous action spaces through Bayesian optimization \cite{morere2018continuous}. Fern{\'a}ndez et al. use partially observable Monte Carlo planning (POMCP) with Gaussian process beliefs for estimating quantiles of the underlying world state \cite{fernandez2022informative}. R{\"u}ckin et al. present a hybrid approach that combines tree search with an offline-learned neural network to predict informative sensing actions \cite{ruckin2022adaptive}.

The AIPPMS problem is less widely studied. Choudhury et al. introduced the AIPPMS problem where the agent must make decisions about where to visit and what sensors to use \cite{choudhury2020adaptive}. Greater energy cost is incurred for using more precise sensors. Choudhury et al. formulate the AIPPMS problem as a POMDP and use POMCP to produce a solution. Ott et al. extend this work by formulating the AIPPMS problem as a belief MDP and use belief-dependent rewards to penalize uncertainty in the belief through sequential Bayesian optimization using Monte Carlo tree search with double progressive widening (MCTS-DPW) \cite{ott2022sequential}. 

While sequential rollout methods have seen great success in application to the AIPPMS problem, they can also suffer from myopia resulting from the limited rollout depth. In trajectories with longer horizons, this myopia can lead to valuable sensing measurements being frontloaded earlier on in the trajectory which can be detrimental to performance depending on the objective function being used. %
In applications such as planetary exploration, frontloading can weaken the performance in characterizing the presence of quantities across the subsurface exploration region \cite{francis2017aegis, mangold2021perseverance}. 

Projection-based trajectory optimization was originally proposed by Hauser as an iterative first order optimization method which allowed the descent direction at each iteration to be calculated using linear quadratic regulator techniques \cite{hauser2002projection}. Miller and Murphey used projection-based trajectory optimization with an ergodic objective based on an underlying expected information density distribution \cite{miller2013trajectory}. An ergodic trajectory is one that spends time in a region proportional to the expected information density distribution \cite{mathew2011metrics}. Dressel and Kochenderfer extended this work and applied it to signal localization on drones \cite{dressel2018optimality, dressel2018efficient}. We directly compare our method with that of Ott et al. and Dressel and Kochenderfer \cite{ott2022sequential, dressel2018optimality}.

\section{Preliminaries}\label{sec:preliminaries}
\subsection{Adaptive Informative Path Planning with Multimodal Sensing} 
The objective in the AIPPMS problem is to determine the optimal sequence of actions, denoted as $\psi^* = (a_1, ..., a_N)$, where each action $a_i$ is a member of a set of potential actions $\mathcal{A}$. Crucially, these actions comprise a hybrid set of both discrete and continuous elements. Discrete elements include choices like the type of sensors used, while continuous elements represent actions such as acceleration commands or future agent positions. The action sequence $\psi^*$ is chosen to maximize an information-theoretic criterion $I(\cdot)$. This can be represented as: \begin{equation} \label{eq:ipp_problem} \psi^* = \argmax_{\psi \in \Psi} I(\psi), \hspace{0.5em} \text{s.t. } C(\psi) \leq B, \end{equation} where $\Psi$ denotes the set of all possible hybrid action sequences of discrete and continuous actions, the cost function $C: \Psi \to \mathbb{R}$ maps these hybrid action sequences to execution costs, $B \in \mathbb{R}^{+}$ is the agent's budget limit, such as time or energy, and $I: \Psi \to \mathbb{R}$ is the information criterion, computed from the new sensor measurements obtained by executing the actions $\psi$ \cite{popovic2020informative, choudhury2020adaptive, ruckin2022adaptive, ott2022sequential}.

\subsection{Gaussian Processes}
We use a Gaussian process to represent the belief over the true environment state. A Gaussian process is a distribution over functions that can be used to predict an underlying function $f$ given some previously observed noisy measurements at location $x$. That is, $y = f(x) + \epsilon$ where $f$ is deterministic but $\epsilon$ is zero-mean Gaussian noise, $\epsilon \sim \mathcal{N}(0, \nu)$. The distribution of functions conditioned on the previously observed measurements  $\hat{y} \mid y, \nu \sim \mathcal{N}(\mu^{\ast}, \Sigma^{\ast})$ is given by: \footnotesize \begin{align} \label{eq:gp_update}
    \mu^{\ast} &= m(X^{\ast})+ K(X^{\ast}, X)(K(X, X) + \nu I)^{-1} (y - m(X)) \\
    \Sigma^{\ast} &= K(X^{\ast}, X^{\ast}) - K(X^{\ast}, X)(K(X,X) + \nu I)^{-1} K(X, X^{\ast})
\end{align} \normalsize where $X$ is the set of measured locations, $X^{\ast}$ is the set of locations to predict the values $\hat{y}$, $m(X)$ is the mean function and $K(X,X')$ is the covariance matrix constructed with the kernel $k(x,x')$ \cite{kochenderfer2019algorithms}.

\subsection{Variance Reduction}
Our approach uses the reduction in variance of the Gaussian process as the exploration objective. Intuitively, this means we want our prediction error about the unseen locations in the environment to be as small as possible.

We denote $\mathcal{X}_{\mathcal{V}}$ as the set of random variables at all locations in the environment and $M_t$ as the locations that have been observed by the agent at time $t$ as well as the corresponding measurement received and sensor type that was used at each of the observed locations. That is, $M_t = \left[x_{0:t}, y_{0:t}, \nu_{0:t} \right]$ where $x_i \in \mathbb{R}^n$ is a point location in the environment, $y_i \in \mathbb{R}$ is the measurement value received and $\nu_i \in \mathbb{R}$ corresponds to the measurement noise in Eq. \eqref{eq:gp_update}. The measurement type is characterized by its level of noise with more precise sensors incurring greater costs. We denote the set of random variables that have been observed by the agent at time $t$ as $\mathcal{X}_{M_{t}}$  which is a subset of all of the random variables at all locations in the environment $\mathcal{X}_{M_{t}} \subseteq \mathcal{X}_{\mathcal{V}}$. The variance reduction of the Gaussian process between the point locations in $M_t$ and the rest of the world $\mathcal{X}_{\mathcal{V}}$ given previous measurement locations and sensor types $M_{t-1}$ is \begin{equation} I(\mathcal{X}_{\mathcal{V} \mid M_t} ; \mathcal{X}_{\mathcal{V} \mid M_{t-1}} ) = \text{Tr} \left( \boldsymbol{\Sigma^{\ast}_{M_{t-1}}} \right) - \text{Tr} \left( \boldsymbol{ \Sigma^{\ast}_{M_{t}}} \right) \end{equation} where $\boldsymbol{ \Sigma^{\ast}_{M_{t}}}$ is the covariance of all points in the environment after measuring at $M_t$ and the trace of the covariance matrix corresponds to the total variance in the world belief-state.

\section{Methods}\label{sec:methods}

\subsection{Projection-based Trajectory Optimization}
The discrete dynamics of a general nonlinear agent can be expressed as $x_{t+1} = x_t + h(x_t,u_t)\Delta t$ where $x_t \in \mathbb{R}^n$ represents the state, $u_t \in \mathbb{R}^m$ the control inputs, and $h$ is the nonlinear dynamics function of the agent. In this work, we focus on AIPPMS problems and use relatively simple dynamics models to allow for more even comparison with other methods; however, our method can be extended to more complex dynamics following the integration methods discussed by Dressel and Kochenderfer \cite{dressel2019tutorial}. The locations that have been observed by the agent at time $T$ where $T$ is the horizon of the trajectory are given by $M_T = \left[x_{0:T}, y_{0:T}, \nu_{0:T} \right]$ and $M_0$ is the empty set. The objective function $J(\cdot)$ is defined as a function of the Gaussian process variance, control effort, and a goal and boundary penalty: %
\begin{equation}
\begin{aligned}
    & J(x, u) = -q\underbrace{I(\mathcal{X}_{M_{T}} ; \mathcal{X}_{\mathcal{V} \mid M_{0}} )}_{\text{variance reduction}} + \underbrace{\lVert x_T - x_f \rVert^2_{Q_f}}_{\text{goal penalty}}  \\ 
    & \hspace{5em} + \underbrace{\sum_0^T \frac{1}{2} u^T_t R_t u_t}_{\text{control effort}} + \underbrace{J_b(x_{0:T})}_{\text{boundary penalty}}
    \label{eq:objective}
\end{aligned}
\end{equation}
\textcolor{black}{where $q \in \mathbb{R}^+$, $R_t \in \mathbb{R}^{m \times m}$, and $Q_f \in \mathbb{R}^{n \times n}$ are arbitrary design parameters that affect the relative importance of minimizing the negative Gaussian process variance, control effort, and distance from the goal location respectively. Note that we include a negative sign on the variance reduction since we seek to minimize $J(x_{0:T}, u_{0:T})$ and therefore maximize the variance reduction. The goal penalty $\lVert x_T - x_f \rVert^2_{Q_f}$ is computed using the Mahalanobis norm with respect to the positive definite matrix $Q_f$. $J_b(x_{0:T})$ is the boundary penalty that penalizes the trajectory for leaving the optimization region as defined by Dressel and Kochenderfer \cite{dressel2019tutorial}. The goal is to solve for the feasible continuous-time trajectory that minimizes the objective function, i.e.} \begin{equation}    
\begin{aligned}
& \underset{x_{0:T},u_{0:T}}{\text{minimize}}
& & J(x_{0:T},u_{0:T}) \\
& \text{subject to} 
& & x_{t+1} = x_t + h(x_t,u_t)\Delta t, \hspace{1em} x_0 = x_{\text{init}}. 
\end{aligned}
\label{eq:general_opt}
\end{equation} The optimization of Equation \ref{eq:general_opt} can be carried out iteratively in the following steps \cite{dressel2019tutorial}.

\subsubsection{Linearize}
\textcolor{black}{To efficiently compute the descent direction, we can linearize the dynamics and control at each state and control input in our current trajectory. That is $A_t = \nabla_x h(x_t, u_t)$ and $B_t = \nabla_u h(x_t, u_t)$. A perturbation of the discrete trajectory is then given by $\delta x_{t+1} = \delta x_t + (A_t \delta x_t + B_t \delta u_t) \Delta t$ where the perturbations $\delta x_t$ and $\delta u_t$ are small changes in the state and control at the time $t$. By introducing $\tilde{A}_t = I + A_t \Delta t $ and $\tilde{B}_t = B_t\Delta t $ we have $\delta x_{t+1} = \tilde{A}_t \delta x_t + \tilde{B}_t \delta u_t$.}

\begin{figure*}[t]
\centering
    {\includegraphics[width=.9\textwidth]{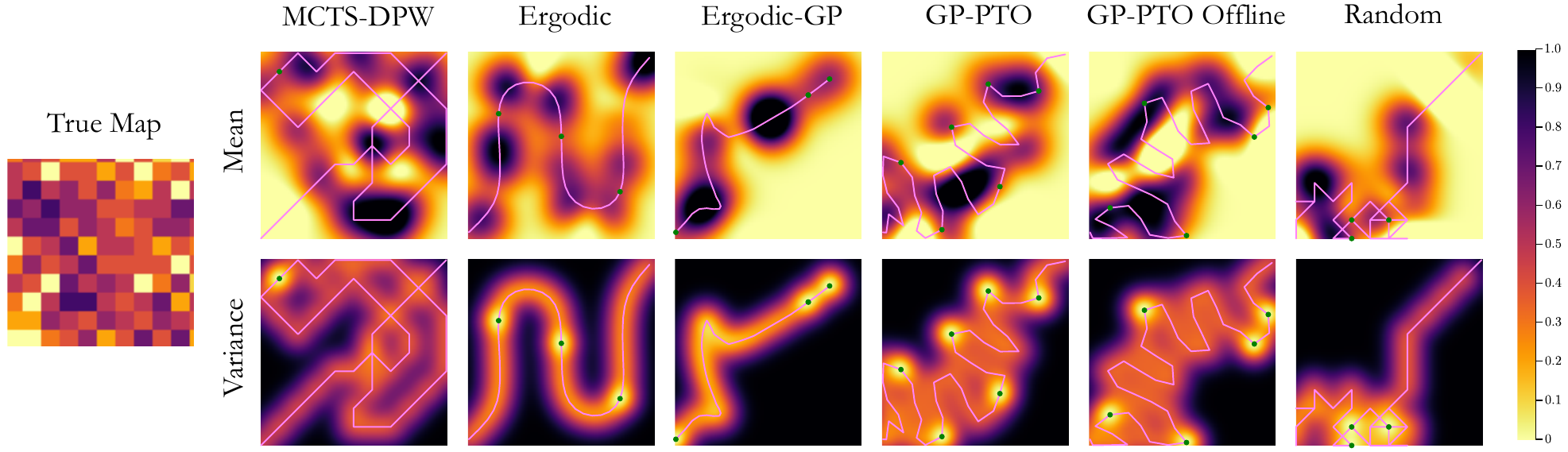}}    
  \caption{Examples of trajectories produced from each of the six methods considered in this work. The true map is shown on the left. The top row shows the posterior mean and the bottom row shows the posterior variance of the Gaussian process world belief after the agent has reached the goal state. The agent starts in the bottom left corner. The pink line indicates the path the agent took and the green dots represent drill sites. These trajectories were generated with $\sigma_s = 1.0$ and $b = 60.0$.} 
  \label{fig:all_trajectories}
  \vspace{-5mm}
\end{figure*}

\subsubsection{Descent Direction}
The trajectory and control descent directions are given by $z_{0:T}$ and $v_{0:T}$ respectively. To compute these descent directions, we can rewrite Equation \ref{eq:general_opt} as shown by Miller and Murphey \cite{miller2013trajectory} as \begin{align*}
& \underset{z_{0:T},v_{0:T}}{\text{minimize}} \sum_0^T \Bigl(a^T_t z_t + b^T_t v_t + \frac{1}{2}z^T_tQ_nz_t + \frac{1}{2} v^T_t R_n v_t \Bigr) \\
& \text{subject to}  \hspace{1em} z_{t+1} = \tilde{A}_tz_t + \tilde{B}_t v_t, \hspace{1em} z_0 = 0. \end{align*} \label{eq:descent_direc}\textcolor{black}{In the above expression, $a_t$ and $b_t$ are the gradients of the objective $J$ with respect to the state and control. That is $a_t = \nabla_x J(x_t,u_t)$ and $b_t = \nabla_u J(x_t,u_t)$. $Q_n$ is an arbitrary positive semi-definite matrices and $R_n$ is positive definite \cite{hauser2002projection}. The initial state descent direction $z_0$ is constrained to zero because the agent's starting state is fixed and cannot be changed. The above expression is a linear quadratic optimal control problem that can be solved using standard discrete algebraic Riccati equations \cite{hauser2002projection, brian1990optimal}. Intuitively, $z_{t}$ and $v_{t}$ tell us how we should modify $x_{t}$ and $u_{t}$.} 

\subsubsection{Candidate Trajectory \& Projection}
\textcolor{black}{Applying the descent direction with step size $\gamma$ will result in a candidate trajectory $\tilde{\alpha}_{0:T}$ and candidate control sequence $\tilde{\mu}_{0:T}$ that may not be dynamically feasible. That is, $\tilde{\alpha}_{0:T} = x_{0:T} + \gamma z_{0:T}$ and $\tilde{\mu}_{0:T} = u_{0:T} + \gamma v_{0:T}$. Therefore, we must project the candidate $(\tilde{\alpha}_{0:T}, \tilde{\mu}_{0:T})$ into a feasible space with the projection operator $\mathscr{P}$:} \begin{equation}
    \mathscr{P}\left(\tilde{\alpha}_{0:T}, \tilde{\mu}_{0:T}, x_0\right):\left\{\begin{array}{l}
    \tilde{u}_t=\tilde{\mu}_t+\mathscr{K}_t\left(\tilde{\alpha}_t -\tilde{x}_t\right) \\
    \tilde{x}_{t+1}= \tilde{x}_t + h\left(\tilde{x}_t, \tilde{u}_t\right) \Delta t
    \end{array}\right.
\end{equation} where $\mathscr{K}_t$ is the Riccati gain computed from the linear quadratic regulator problem with respect to the reference candidate trajectory \cite{miller2013trajectory, dressel2019tutorial}. A high-level overview of this approach is shown in Fig. \ref{fig:front_page_fig}. For implementation details, we refer the reader to our open-source repository. %

Previous projection-based trajectory optimization methods have focused solely on ergodic objectives, while our contribution distinctively expands the scope to Gaussian process-based objectives and incorporates multimodal sensor selection \cite{miller2013trajectory, dressel2019tutorial}. A comparative analysis of these methodologies is presented in the subsequent section.

\subsection{Sample Selection}
\textcolor{black}{The projection-based trajectory optimization controls how the shape of the trajectory evolves, but the variance reduction term in Equation \ref{eq:objective} is also dependent on the sample type. Therefore, during the iterative optimization, we perturb one sensor type $\nu_i$ in $M_t = \left[x_{0:t}, y_{0:t}, \nu_{0:t} \right]$ with probability $p$ where $i$ is chosen from a uniform distribution at each iteration. If the overall objective decreased, then the perturbation is kept, otherwise, it is discarded. Note that this random perturbation allows for both continuous and discrete sensor types since the effect of continuous sensor types can be analyzed analytically and discrete sensor types can be perturbed randomly by uniformly sampling an integer from the range of sample types available. Due to energy constraints, inserting a more precise sensor type will lead to the trajectory being shortened since more energy resources will be needed for that specific sensor type. As a result, inserting valuable sensor types will not always necessarily lead to a decrease in the objective function. Previous projection-based trajectory optimization approaches have not incorporated multimodal sensor selection.}

\section{Results}\label{sec:results}
\begin{figure*}[t]
\centering
    {\includegraphics[width=1.0\textwidth]{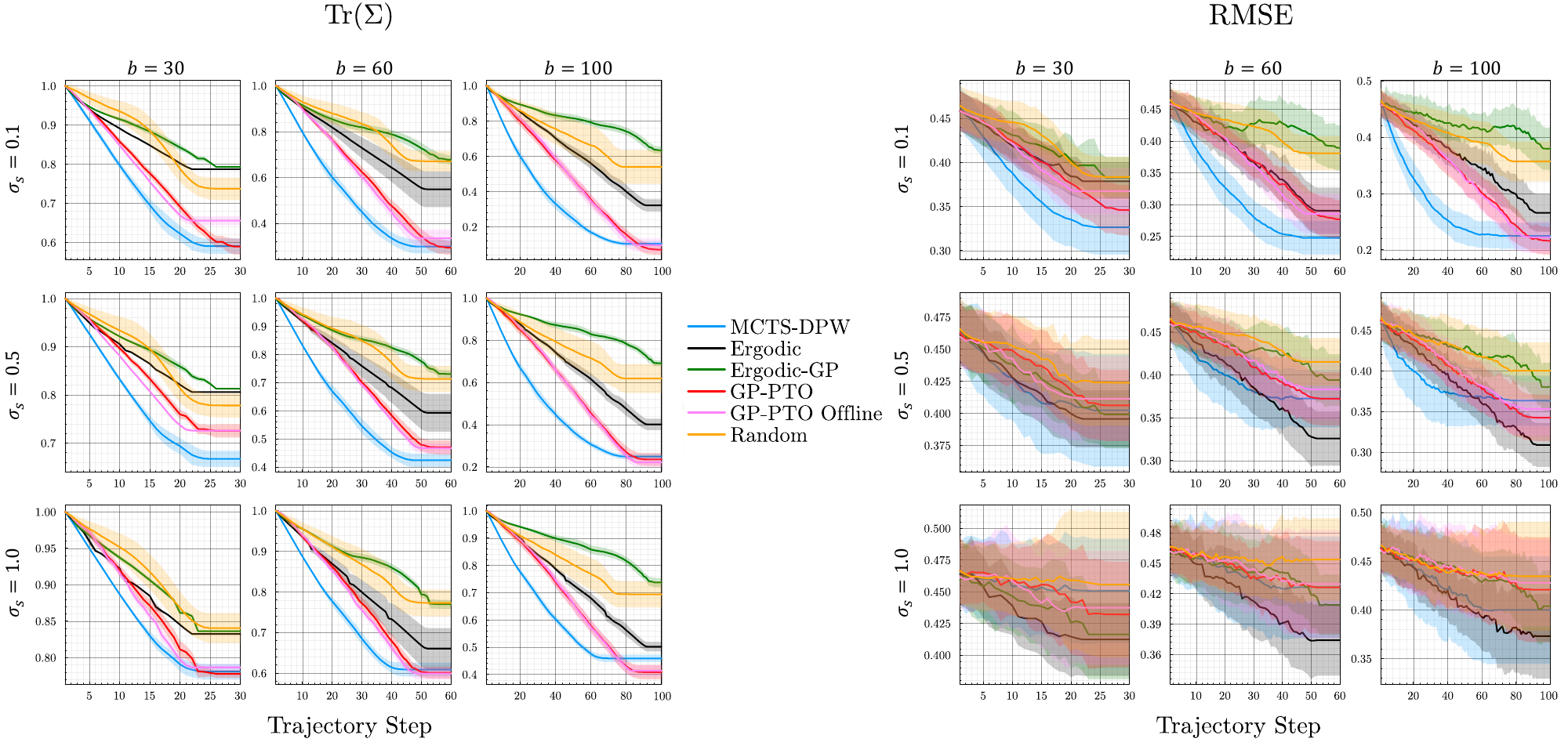}}
  \caption{The trace of the Gaussian process covariance matrix is shown on the left and the RMSE of the beliefs with respect to the true map is shown on the right. For even comparison we evaluated $\text{Tr}(\Sigma)$ and the RMSE all using the same Gaussian process setup, even though the Ergodic and Random methods do not use the Gaussian process for decision making. Each subplot shows the average and standard deviation from 50 simulation runs. The budget was varied from $b=30$ to $100$ and the spectrometer noise was varied from $\sigma_s = 0.1$ to $1.0$.} 
  \label{fig:variance_rmse}
  \vspace{-6mm}
\end{figure*}

\subsection{Rover Exploration Problem}
To evaluate our approach, we use a modified version of the Rover Exploration problem which was a benchmark AIPPMS problem introduced by Ott et al. and inspired by multiple planetary rover exploration missions \cite{francis2017aegis, mangold2021perseverance, heldmann2016site, ott2022sequential}. The rover begins at a specified starting location and has a fixed amount of energy available to explore the environment and reach the goal location. The rover is equipped with a spectrometer and a drill. Drilling reveals the true state of the environment at the location the drill sample was taken and is a more costly action to take from a resource budget perspective. Conversely, the spectrometer provides a noisy observation of the environment and uses less of the resource budget. Since the rover does not have access to the ground truth state of the environment during execution, the rover's goal is to reduce the uncertainty in its belief about the environment as much as possible.%

The environment is modeled as an $n \times n$ grid with $\beta$ unique measurement types in the environment. To construct a spatially-correlated environment, we first start by sampling each grid cell value from an independent and identically distributed uniform distribution of the $\beta$ unique measurement types. Then, each cell takes on the average value of its neighboring cells with probability $p_g$. This process creates environments with spatial correlation as well as some random noise to simulate geospatial environments where the amount of spatial correlation is controlled by $p_g$.

\subsection{Approaches Considered}
We directly compare our approach with five other methods.
\begin{enumerate}[leftmargin=5mm]
\item \textbf{MCTS-DPW}. Ott et al. used MCTS-DPW with Gaussian process beliefs which was able to significantly outperform previous AIPPMS approaches \cite{ott2022sequential, choudhury2020adaptive}.   
\item \textbf{Ergodic}. Projection-based trajectory optimization has previously been applied to ergodic objectives introduced by Miller and adapted by Dressel \cite{miller2013trajectory, dressel2018optimality}. We use evenly spaced drill measurements since the ergodic metric does not consider multimodal sensing capabilities.
\item \textbf{Ergodic-GP}. The Ergodic Gaussian process (Ergodic-GP) method is a modified ergodic control approach that also uses a Gaussian process to update the information distribution as samples are received online. The updated information distribution is then used to plan an ergodic trajectory at the next iteration. This approach allows the agent to combine the ergodic metric for trajectory generation with the Gaussian process for drill site selection.
\item \textbf{GP-PTO}. The Gaussian Process Projection-based Trajectory Optimization (GP-PTO) method is our contribution outlined in the previous section. 
\item \textbf{GP-PTO Offline}. The GP-PTO Offline method is a slight modification of GP-PTO that allows the optimization to run longer and then executes the best trajectory without replanning online. 
\item \textbf{Random}. The random method was implemented as a baseline. At each step, the agent chooses randomly from a set of actions. The action space is constrained so that the agent will always be able to reach the goal location within the specified energy resource constraints. 
\end{enumerate}

MCTS-DPW, GP-PTO, Ergodic-GP and Random all replan after each step of the trajectory. This means that they receive a sample, update their belief, and then replan accounting for this updated belief. The Ergodic and GP-PTO Offline methods do all of their planning offline and then execute the trajectory without accounting for the sample types that were received.

\subsection{Trajectory Comparison}
An example of trajectories generated with each of the six approaches is shown in Fig. \ref{fig:all_trajectories}. We highlight a few key observations that emerge from these trajectories. MCTS-DPW often revisits locations in the environment that it has already explored. MCTS-DPW uses rollouts to estimate the value of each action available to it. As a result, exploration is limited to its rollout depth and therefore it cannot take into consideration the entire trajectory during planning, but instead only considers the next several steps in the trajectory. %

The Ergodic approach produces a sinusoidal trajectory, which is often desirable in exploration and search-based problems \cite{mathew2011metrics, miller2013trajectory, dressel2018optimality}. However, since the ergodic metric has no knowledge of the underlying Gaussian process, it is not able to adjust the spacing of these sinusoidal waves accordingly, which leaves many gaps unexplored. 

We see relatively poor results from the Ergodic-GP approach. %
The issue arises due to oscillations in the ergodic metric. In the Fourier decomposition method used by Miller and Dressel, the trajectories can have significant changes in direction from one planning iteration to the next, causing oscillations \cite{miller2013trajectory, dressel2018optimality}. While the trajectories planned at each iteration are feasible, what ends up being executed is often of poor quality. This poor performance under rapid replanning has also been noted by Dressel \cite{dressel2018efficient}. 

The GP-PTO and GP-PTO Offline methods are able to smartly place drill sites on the outer regions of the trajectory. This emergent behavior results from the fact that the GP-PTO methods recognize they will not be able to visit the entire environment so they dedicate their most valuable samples to the boundaries and accept a greater quantity of noisier samples on the interior. \textcolor{black}{Placing valuable measurement types at the boundary of the explored region will decrease the variance more compared to placing the valuable measurements on the inner regions of previously explored space.}

\subsection{Simulation Results}
\begin{figure}[t]
\centering
    {\includegraphics[width=1\columnwidth]{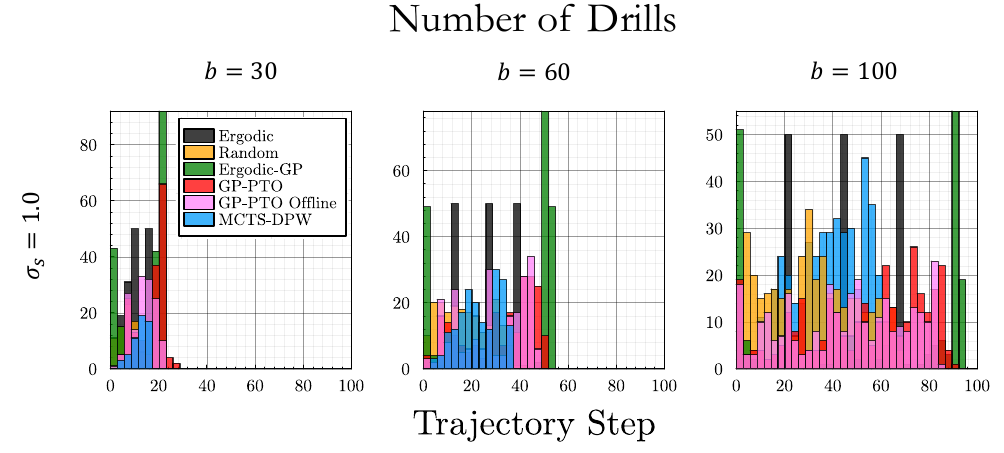}}    
  \caption{Drill placements along the trajectory from 50 simulations.} \label{fig:drills}
  \vspace{-5mm}
\end{figure}

For the Rover Exploration problem, we focus on the interplay between the resource budget allotted to the agent and the sensing quality of the spectrometer, where $\sigma_s$ denotes the standard deviation of a Gaussian sensor model. The results from our large scale simulation comparison across varying $\sigma_s$ and $b$ are summarized in Figs. \ref{fig:variance_rmse} and \ref{fig:drills}. 

GP-PTO outperforms MCTS-DPW in total variance reduction in roughly four of the nine cases considered. They approximately tie in three cases and MCTS-DPW outperforms in two cases. Of particular significance is that GP-PTO tends to outperform as the resource budget is increased. GP-PTO considers the entire trajectory during the optimization, while MCTS-DPW only rolls out a few steps into the future. Therefore, as the resource budget is increased, accounting for future actions becomes more important to performance. MCTS-DPW consistently has a steeper slope in variance reduction which is a direct result of the differences between the rollout-based methods and projection-based methods. MCTS-DPW will take the actions that have the greatest variance reduction based on its rollout horizon and therefore will prioritize more near-term variance reduction whereas the GP-PTO methods will prioritize the long-term variance reduction. In trajectories with less energy budget, MCTS-DPW does not have an issue because there are fewer opportunities to re-explore areas it already visited. However, in longer trajectories, MCTS-DPW overlaps itself much more frequently as shown in Fig. \ref{fig:all_trajectories} due to its limited rollout depth. Additionally, MCTS-DPW uses a discrete action space so all of its movements are of unit size. Conversely, the projection-based methods have a hybrid action space and therefore reason over variable step sizes. 

The GP-PTO methods perform relatively similarly, and in one case, the online version outperforms the offline version. GP-PTO offline was allowed to run for $5,000$ iterations whereas GP-PTO ran until convergence criteria were met or a maximum of $50$ iterations was reached. Additionally, the GP-PTO method can be terminated at any point based on planning time, iterations, or other convergence criteria and the best solution found up to that point will be used. 

The Ergodic approach is clearly inferior but still performs relatively well in comparison with the other methods. This is a very interesting result since the Ergodic method has no knowledge of the Gaussian process world belief, yet is still able to perform well indicating that there are some similarities present between the objective of reducing the variance of a Gaussian process and the measure of ergodicity. %
The Ergodic-GP approach is clearly the worst performer -- sometimes even being outperformed by the Random policy. %

Fig. \ref{fig:drills} shows the distribution of drill sites along all of the trajectories from the simulation runs.  We can see that when the spectrometer produces very noisy measurements, $\sigma_s = 1.0$, MCTS-DPW and GP-PTO tend to drill more frequently with the drills distributed throughout the trajectory. 

It is also important to note the differences in planning time across the different approaches considered. In all of our experiments, the average planning time for MCTS-DPW, Ergodic, Ergodic-GP, GP-PTO, and GP-PTO Offline were on the order of $0.50, 0.001, 3.0, 10.0, 30.0$ seconds respectively. Note that the increased planning time in the GP-PTO methods is a result of conducting inference over the Gaussian process belief for the whole trajectory. This is the main bottleneck in the planning process, but additional planning time constraints can be imposed and the GP-PTO methods will stop execution and return the best solution found so far. %

\subsection{Expected Improvement Metric}
\begin{figure}[t]
\centering
    {\includegraphics[width=1.0\columnwidth]{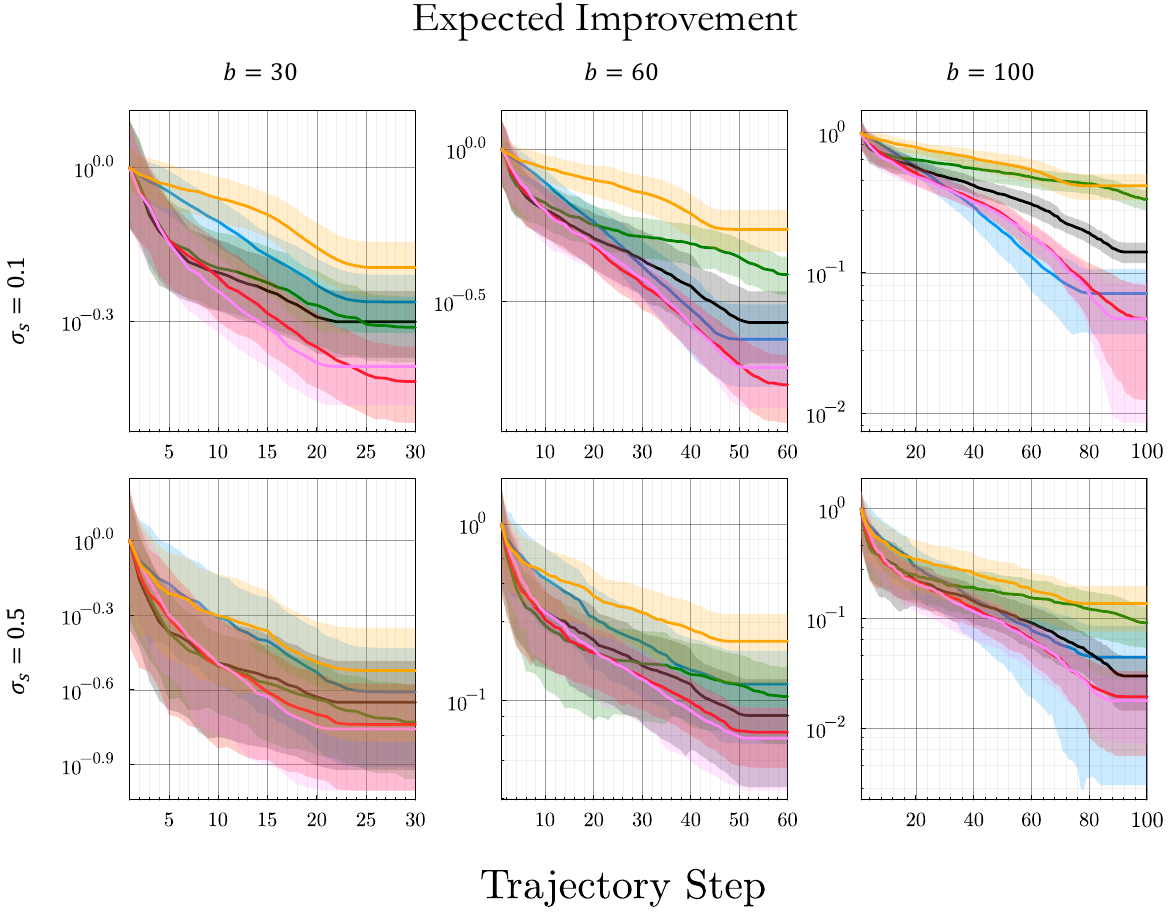}}
  \caption{Expected improvement results with respect to the true map. The average and standard deviation from 50 simulation runs are shown.} 
  \label{fig:ae_expected_improvement}
  \vspace{-6mm}
\end{figure}

The predictive variance in Gaussian processes depends only on the locations of measurements. We have also evaluated our method using the expected improvement metric which does take sample values into account. These results are shown in Figure \ref{fig:ae_expected_improvement}. The expected improvement metric is given by: $$\mathbb{E}[I_p(y)] = (y_{\text{min}} - \hat{\mu}) P(y \leq y_{\text{min}}) + \hat{\sigma}^2 \mathcal{N}(y_{\text{min}} \mid \hat{\mu}, \hat{\sigma}^2) $$ where $I_p(y)$ is the improvement over the current minimum found so far, $y_{\text{min}}$ is the minimum value sampled so far, and $\hat{\mu}$ and $\hat{\sigma}^2$ are the predicted mean and variance of the Gaussian process \cite{kochenderfer2019algorithms}. We see that GP-PTO is also able to outperform the other methods considered on this adaptive metric.

\section{Conclusion}
This work focused on a variant of the adaptive informative path planning problem where the agent is equipped with multiple sensors. We introduced a Gaussian process projection-based trajectory optimization approach where we considered both sensor distribution and sensor type. We showed that our approach demonstrates performance comparable to previous state-of-the-art AIPPMS methods, but offers scalability advantages in long horizon trajectories by reasoning about the entire trajectory during the optimization process. Future work includes integrating more efficient Gaussian process representations to decrease planning time, such as adaptive query resolutions and
expanding the belief representation to reason over Gaussian
process parameters. Additionally, alternative sample injection techniques and objective functions can be explored for improved performance. Lastly, hybrid approaches have the potential to present unique solutions. For example, combining MCTS-DPW with GP-PTO could result in more efficient solutions. %
These areas of future research would allow a similar formulation to be extended to a variety of real-world problems.

\printbibliography

\end{document}